\documentclass[a4paper, conference]{IEEEtran}  

\IEEEoverridecommandlockouts                              

\usepackage{amsmath}
\usepackage{xfrac}
\usepackage{physics}
\usepackage{xcolor}
\usepackage{multirow}
\usepackage{hyperref}
\hypersetup{
  pdftitle={A Biomimetic Fingerprint for Robotic Tactile Sensing},
  pdfauthor={Oscar Alberto Jui\~na Quilachamin, and Nicolás Navarro-Guerrero},
  pdfsubject={Robotics (cs.RO); Artificial Intelligence (cs.AI),
  56th International Symposium on Robotics (ISR Europe 2023)},%
  pdfkeywords={tactile sensing, dataset, robot perception},%
  colorlinks=true,
  linktoc=all,
  linkcolor=darkgray,
  citecolor=darkgray,
  urlcolor=black,
}

\usepackage[protrusion=true,expansion=true]{microtype}

\begin{document}
\title{A Biomimetic Fingerprint for Robotic Tactile Sensing}

\author{\IEEEauthorblockN{Oscar Alberto Jui\~na Quilachamin}
\IEEEauthorblockA{\textit{Faculty of Mechatronic \& Micro-Mechatronic Systems}\\
\textit{Hochschule Karlsruhe (HKA)}\\
Karlsruhe, Germany\\
oscar.juina@eu4m.eu}
\and
\IEEEauthorblockN{Nicol\'as Navarro-Guerrero}
\IEEEauthorblockA{\textit{Deutsches Forschungszentrum für Künstliche Intelligenz GmbH}\\
\textit{and, L3S Research Center, Leibniz Universität Hannover}\\
Hannover, Germany\\
\url{https://orcid.org/0000-0003-1164-5579}}
}


\maketitle

\begin{abstract}
Tactile sensors have been developed since the early '70s and have greatly improved, but there are still no widely adopted solutions. Various technologies, such as capacitive, piezoelectric, piezoresistive, optical, and magnetic, are used in haptic sensing. However, most sensors are not mechanically robust for many applications and cannot cope well with curved or sizeable surfaces. 
Aiming to address this problem, we present a 3D-printed fingerprint pattern to enhance the body-borne vibration signal for dynamic tactile feedback. 
The 3D-printed fingerprint patterns were designed and tested for an RH8D Adult size Robot Hand. The patterns significantly increased the signal's power to over 11 times the baseline. 
A public haptic dataset including 52 objects of several materials was created using the best fingerprint pattern and material.
\end{abstract}

\section{Introduction}
Tactile perception is a pivotal technology to enable applications such as dexterous robotic grasping, smart prostheses, and surgical robots \cite{Navarro-Guerrero2023VisuoHaptic}. 
Tactile sensors are mainly designed to mimic mechanoreceptors. Some of the objectives of tactile sensors are to determine the location, shape and intensity of contacts \cite{Pestell2022Artificial}. 
Instantaneous pressure or force can be used to determine the force and multiple contact points. 
In contrast, dynamic tactile sensations are better suited to extract information about texture as well as rolling or slipping \cite{Howe1993Dynamic}. 

Several technologies have been studied over the years \cite{Chi2018Recent}, including capacitive (e.g., \cite{Larson2016Highly}), piezoelectric (e.g., \cite{Seminara2013Piezoelectric}), piezoresistive (e.g., \cite{Jung2015Piezoresistive}), optical (e.g., \cite{Kuppuswamy2020SoftBubble,Ward-Cherrier2018TacTip}), fiber optics (e.g., \cite{Polygerinos2010MRICompatible}), and magnetic (e.g., \cite{Jamone2015Highly}).
While experimental sensors cover many technologies, commercial solutions are less diverse. The most common commercial sensors are those based on optics/cameras (e.g., the PapillArray by Contactile or DIGIT by GelSight), magnetic (e.g., uSkin sensor by Xela Robotics) or piezoelectric (e.g., FTS Tactile Pressure Sensors by Seed Robotics and BioTac\textsuperscript{\textregistered} sensor by SynTouch\textsuperscript{\textregistered}). For more information about haptic perception see \cite{Navarro-Guerrero2023VisuoHaptic, Chi2018Recent}.

\begin{figure}[tb]
  \centering
  \includegraphics[width=0.49\columnwidth]{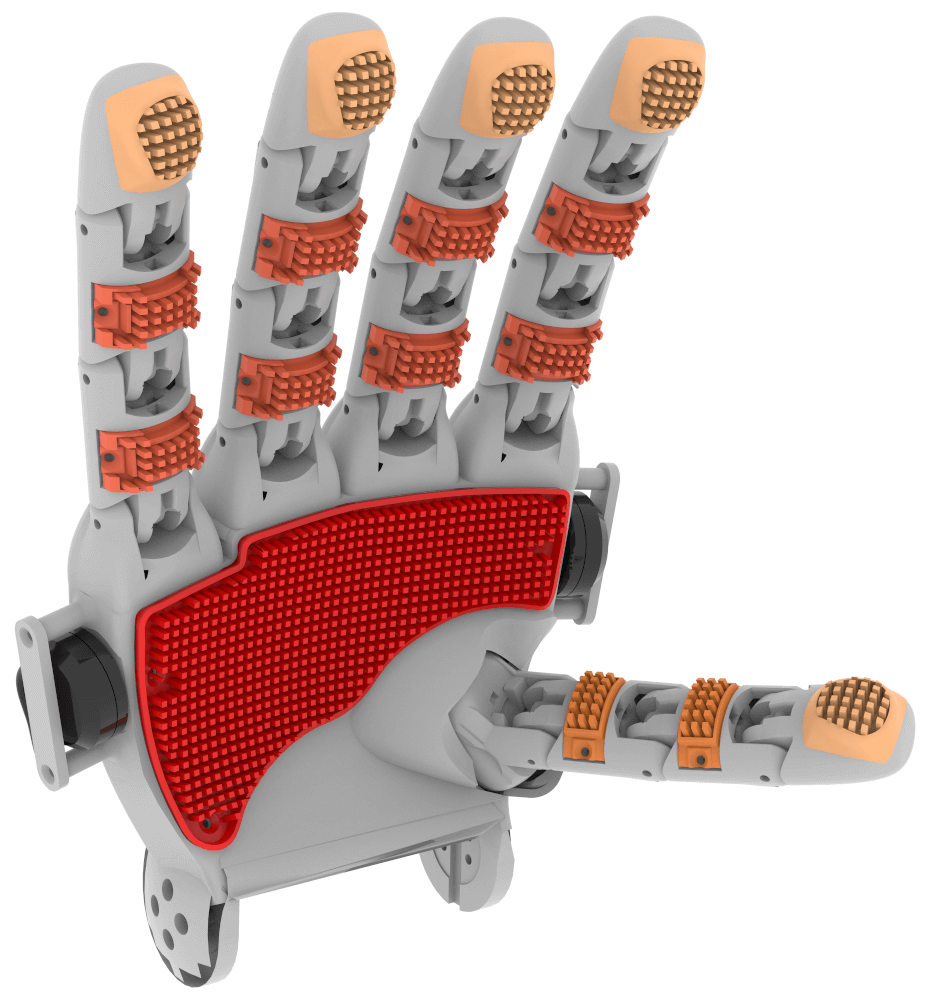}
  \includegraphics[width=0.49\columnwidth]{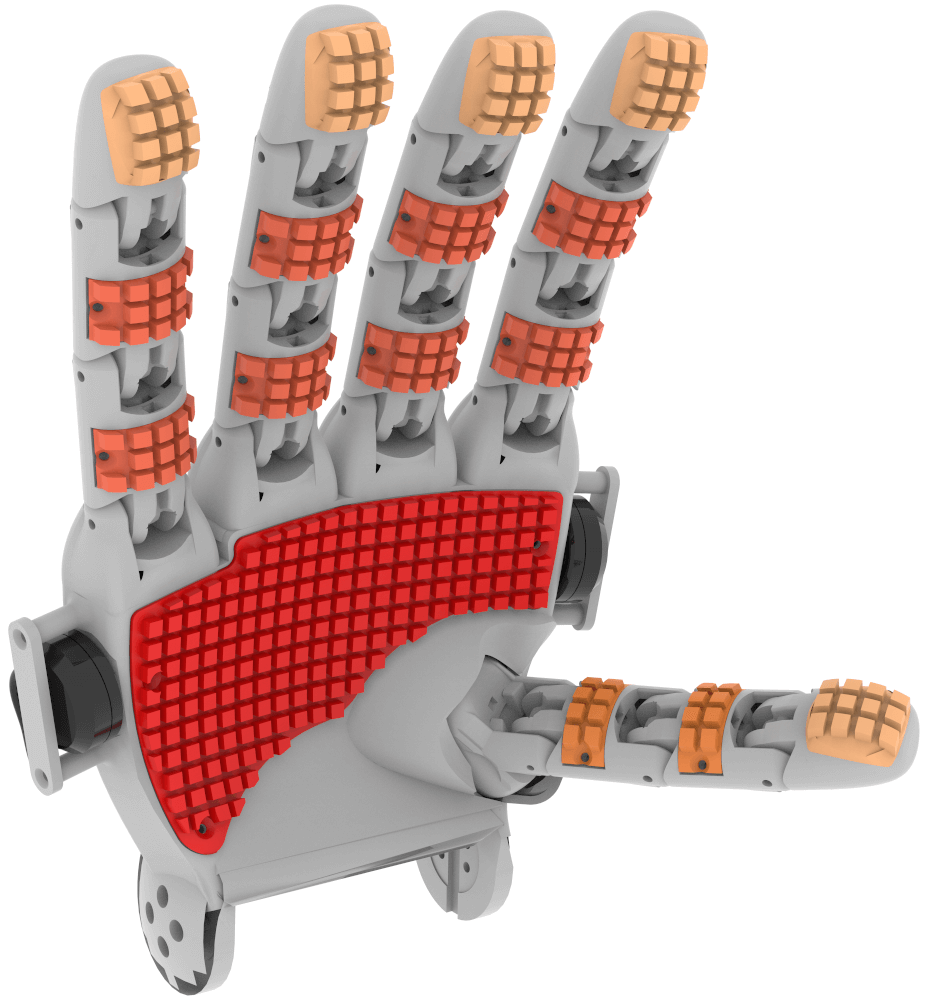}
  \caption{RH8D hand with fingerprint patterns. Left: ST 45B resin. Right: TPU.}
  \label{fig:RH8D-fingerprints}
\end{figure}

\begin{figure}[htbp]
  \centering
  \includegraphics[width=\columnwidth]{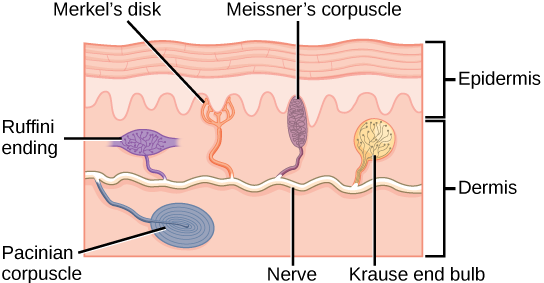}
  \caption{Primary mechanoreceptors in the human skin. Merkel's cells respond to light touch, Meissner's corpuscles respond to touch and low-frequency vibrations. Rufinni endings respond to deformations and warmth. Pacinian corpuscles respond to transient pressure and high-frequency vibrations. Krause end bulbs respond to cold. Image from \cite{Clark2020Biology} CC BY 4.0.}
  \label{fig:skin}
\end{figure}

Inspired by some of the primary mechanoreceptors in the human skin (Fig.~\ref{fig:skin}) and existing related work, we suggest exploiting body-borne vibrations for dynamic tactile perception, which could be used along with other tactile sensors. In particular, we propose 3D-printed fingerprints to increase the body-borne vibrations and signal-to-noise ratio generated when interacting with objects and surfaces. In this version, we used a symmetric beam array covering most of the inner side of the hand, as shown in Fig.~\ref{fig:RH8D-fingerprints}. We optimized the beams' geometry to increase the vibration signals' spectral magnitude within the audible range, see Section \ref{sec:theoretical-analysis}. The vibrations are measured with contact microphones mounted on either side of the robot's palm and one inside the palm. 
Our results show that the optimized 3D-printed fingerprint patterns achieve over 11 times higher spectral magnitude than non-optimized fingerprints. However, additional research is needed to design patterns capable of dealing with various object materials and textures. Moreover, research on the effect of wear, angle of incidence, force and velocity of the beam is also needed.

In a future iteration, the microphones would be mounted inside the robot and thus protected from the elements, collisions, and other perturbations. Additionally, the 3D-printed fingerprint patterns are easy to manufacture, can cover sizeable and curved surfaces.

The paper is organized as follows. Section \ref{sec:sota} presents the related work. Section \ref{sec:requirements} outlines the system requirements and constraints. Section \ref{sec:theoretical-analysis} describes the theoretical analysis and fingerprint design. 
Section \ref{sec:results} presents the results. Section \ref{sec:discussion} discusses the results, and Section \ref{sec:conclusions} presents conclusions and future work.

\section{Related Work}
\label{sec:sota}
Over three decades ago, vibrotactile sensing for robot perception was suggested (e.g., \cite{Howe1989Sensing}). 
One of the first sensors consisted of a core enveloped by polyurethane foam and a stiffer textured outer skin of rubber. An accelerometer was mounted on the inner side of the skin to measure the vibration produced when the finger made or broke contact with an object, an object was lifted or placed on a tabletop, or slippage occurred \cite{Howe1990Grasping}.
The outer skin was textured because very smooth rubber has a substantial coefficient of friction leading to a stick-slip motion of the finger, which in turn, made manipulation more complicated and could overload the accelerometer \cite{Howe1989Sensing}. 
The textured skin consisted of parallel ridges a few hundred microns wide and up to a few hundred microns high. 
Such a pattern causes a ``catch and snap back'' behaviour, leading to local accelerations of the skin, which can be easily detected by the accelerometer \cite{Howe1989Sensing}.

Another version using a similar construction and sensing principles using piezoelectric film strips, which can transduce stress into charge, was later suggested \cite{Howe1993Dynamic}.
The transduction function between stress and charge changes depending on the direction of the stress but is more significant in the transverse direction of the film \cite{Son1994Tactile}.
Hence, the tested sensor consists of 4 strips oriented along the direction of the expected slip for maximum sensitivity \cite{Son1994Tactile}.
Additionally, the skin was textured with protruding nibs to maximize the stress.  

More recently, microphones have been suggested as tactile sensors. 
For instance, Edwards et al.\ \cite{Edwards2008Extracting} presented a setup of a finger with latex skin and a microphone mounted roughly at the position of the fingernail. The untextured latex skin could produce and transfer enough vibrations to the microphone when interacting with artificial symmetrical texture patterns.

Similarly, Hughes et al.\ \cite{Hughes2014Soft} suggested a soft, amorphous texture-sensitive skin based on a sensor network of microphones. The sensors are mounted $15 cm$ apart on a ﬂexible neoprene rubber mesh and then embedded into silicone rubber textured with a 60-grit (ca.\ $0.25 mm$) aluminium oxide sandpaper. 
The sensor spacing corresponds to the wavelength of sound at $250 Hz$, which is considered the mid-frequency of the Pacinian corpuscle \cite{Hughes2014Soft}.
Similarly, the grit size used is considered the groove width of human fingerprints.
A silicone rubber with Young's modulus of $125 kPa$ was used; by reference, the human skin's modulus ranges from 420 to 850 $kPa$.
The skin assembly was systematically tested using a vibration motor. Texture recognition and sound localization show satisfactory results.

Yang et al. \cite{Yang2021Large} suggested an array of condenser microphones as a large dynamic tactile sensor.
The proposed system comprises a porous structured mesh, neoprene, and fabric to form a sensor's skin. The fabric's rough texture generates vibrations that are then measured by the microphones. The neoprene rubber keeps the fabric in place. At the same time, the mesh helps to keep the sensor's shape and transfer the vibrations from the skin to the microphones. 
Time Difference of Arrival (TDoA) was used to determine the point of contact, while a CNN was used to classify the touch type.
The suggested solution can be arbitrarily large and cope well with curvatures and orientations \cite{Yang2021Large}.

Microphones have also been suggested for texture detection in prosthetics \cite{Svensson2021Electrotactile}. In particular, the vibrations a microphone captures while stroking materials of different textures are translated into electrotactile feedback. Despite using an off-the-shelf microphone without an artificial skin to enhance the vibrations, the setup conveys sufficient information to the human such that participants can discriminate textures with 85\% accuracy \cite{Svensson2021Electrotactile}.

Fingerprint-like textured skin has also been used with capacitive and piezoelectric to improve static and dynamic tactile sensing, e.g., ridges \cite{Navaraj2019FingerprintEnhanced, Toprak2018Evaluating} and nibs \cite{Bonner2021AU}.
In particular, Navaraj et al.\ \cite{Navaraj2019FingerprintEnhanced} the 3D printed pattern included ridges of $500 \mu m$ width, $2 mm$ length and $500 \mu m$ thickness. The ridges' vertical separation was $1500 \mu m$ ($>3$ times higher than typically observed in human ﬁngerprints), and horizontal separation of 3000 $\mu m$ \cite{Navaraj2019FingerprintEnhanced}.
The suggested setup reached a maximum accuracy of 99.45\% \cite{Navaraj2019FingerprintEnhanced}.

Pestell et al. \cite{Pestell2022Artificiala} suggested a biomimetic tactile sensor called the TacTip. The sensor aims to mimic human skin's shallow dermal and epidermal layers. The `epidermis' was made from a rubber-like material over a soft inner elastomer gel analogue to the `dermis'. 
These two materials are interdigitated in a mesh of ridges comprising stiff inner nodular pins that extend from the epidermis into the soft gel. Such structure mechanically amplifies skin surface deformation into a lateral movement of the pins, which can be optically tracked and classified \cite{Pestell2022Artificiala}. 
The same structure can be used to detect the induced vibration in the soft inner gel \cite{Pestell2022Artificial}. The fingerprint amplifies the vibrations aiding texture perception. Moreover, the harmonic structure of induced vibrations seems to be speed-invariant \cite{Pestell2022Artificial}.

As shown in this section, vibration, textured skins, and microphones for dynamic tactile sensing have been validated in different setups. 
However, the design decisions for the fingerprint pattern are not clear. In most cases, the aim is to resemble human fingerprints disregarding the different mechanical properties of the artificial skin or transductors.
Thus, we present an approach to optimize 3D-printed fingerprints for dynamic tactile sensing to increase the vibration signal's spectral magnitude and improve the signal-to-noise ratio. We first describe the requirements and constraints and then continue with the theoretical analysis and design. 

\section{System Requirements and Constraints}
\label{sec:requirements}
\subsection{The Hand}
A RH8D robotic hand by \href{https://www.seedrobotics.com/}{Seed Robotics} was used, see Fig.~\ref{fig:RH8D-fingerprints}. The customization corresponds to the microphone holders on either side of the palm and one inside the palm, and a replaceable palm and finger segments marked in hues of red in Fig.~\ref{fig:RH8D-fingerprints}.
The manufacturer provides no detailed specifications for force and speed of the hand.
We estimated the maximum linear velocity and maximum forces generated by the RH8D robot hand to be $\hat{V} = 953.3 \sfrac{mm}{s}$ at $134.5 Nmm$ and $\hat{F} = 473.5 N mm$ at $270.7 \sfrac{mm}{s}$ based on the maximum speed ($78 rpm$) and shall torque ($10 N m$) of the Dynamixel MX motor series and the cable pulley transmission of the RH8D hand.
The estimated weight of each finger assembly is $10.9 g$ and $8.9 g$ for the thumb.

\subsection{The 3D Printer}
We tested two kinds of 3D printer systems, i.e., Fused Deposition Modeling (FDM) and Stereolithography (SLA). We tested PLA ($\rho = 1.17$ to $1.24 \sfrac{g}{cm^3}$ and $E = 2641 MPa$), and TPU ($\rho = 1.22 \sfrac{g}{cm^3}$ and $E = 9 MPa$) for the FDM system and the ST 45B resin ($\rho = 1.20 \sfrac{g}{cm^3}$ and $E = 2000 MPa$) for the SLA system.
Although 3D printing allows for creating custom intricated shapes, the 3D printing technology and printer can restrict the size and shapes that can be created. In particular, we adhere to the following guidelines \cite{Xometry2020Stereolithography}: minimum beam width $\geq 0.4 mm$ for supported printing or $\geq 0.6 mm$ for unsupported printing. Holes should have a diameter $\geq 0.75 mm$.

\subsection{The Sensors}
\begin{figure}[htb]
  \centering
  \includegraphics[width=\columnwidth]{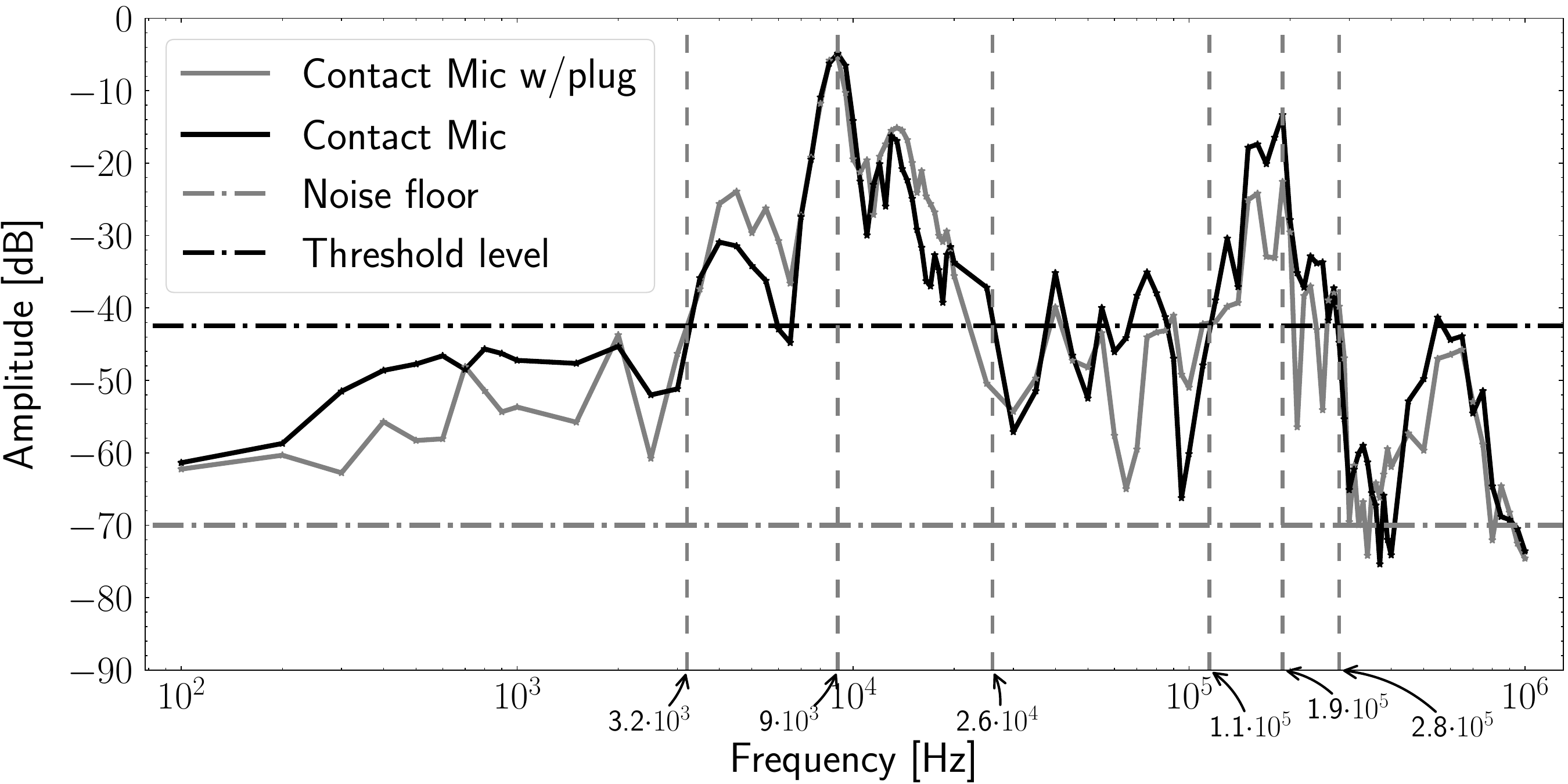}
  \caption{Measured frequency response of a contact microphone with the audio jack (plug) and without. Average noise floor of $-70 dB$. Amplitude threshold of $-42 dB$ corresponding to the mean value of the measured amplitude range.}
  \label{fig:frequency-response-microphones}
\end{figure}

\begin{figure}[htb]
  \centering
  \includegraphics[width=\columnwidth]{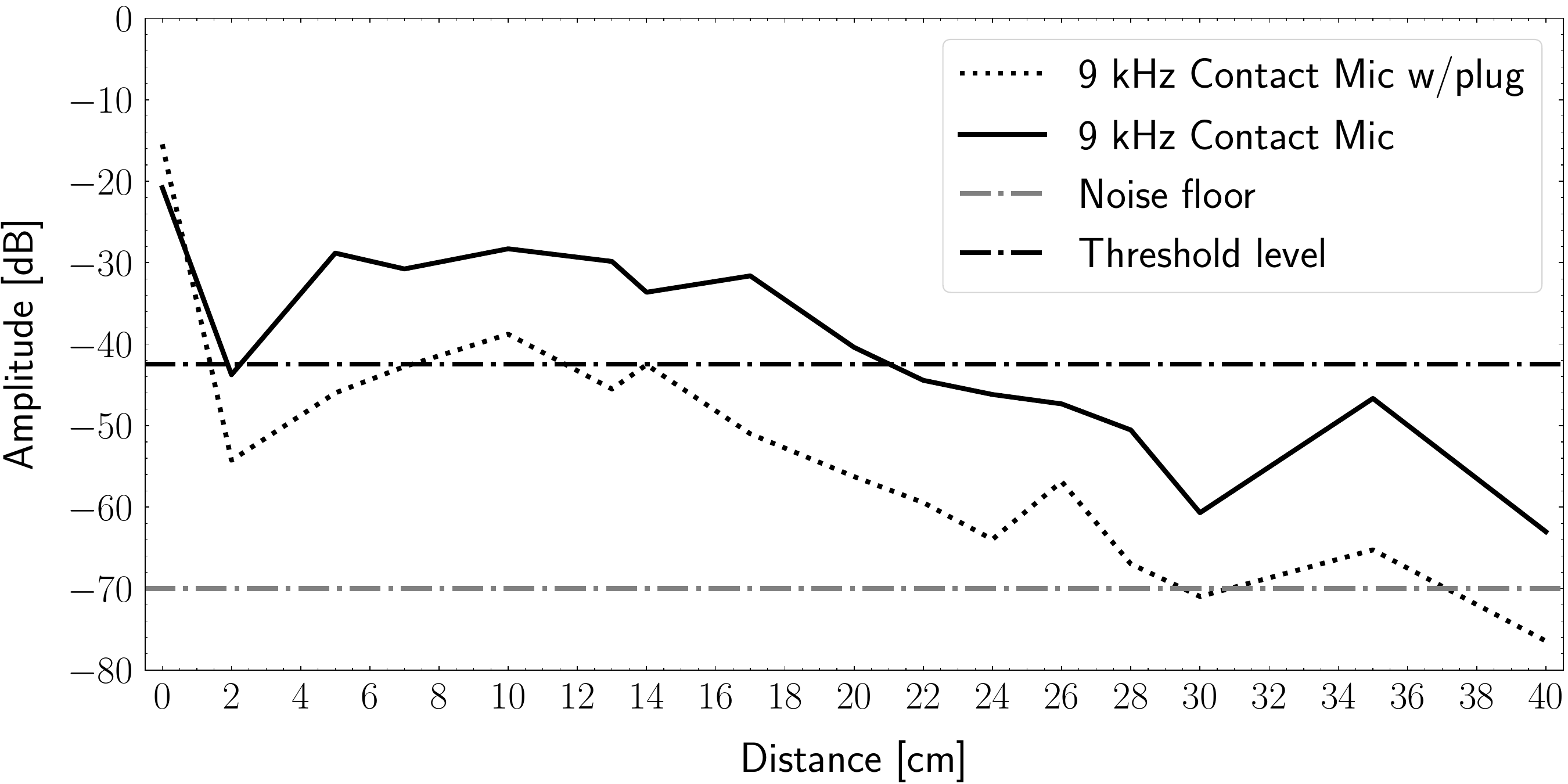}
  \caption{Measured frequency attenuation with respect to distance for $9 kHz$ measured on a solid wood ruler. Measurements for a microphone with the audio jack (plug) and one without.}
  \label{fig:frequency-attenuation-distance}
\end{figure}

Harley Benton CM-1000 contact microphones were used to collect the sound propagated within the robot hand. The microphone is inexpensive and widely available, but no public datasheet exists. Thus, we first estimate some of its relevant characteristics. 
The measured frequency response of the microphones is shown in Fig.~\ref{fig:frequency-response-microphones} and the attenuation with respect to distance for the frequency of highest sensitivity is shown in Fig.~\ref{fig:frequency-attenuation-distance}. 
A minimum desirable amplitude threshold of $-42 dB$ corresponding to the mean value of the measured amplitude range was selected. 
The target frequency ranges were defined based on the measured frequency response and minimum desirable amplitude threshold. 
The first range is between $[3.2, 26] kHz$ with a peak at $9 kHz$ and another between $[110, 280] kHz$ with a peak at $150 kHz$.
The lower frequency range was selected to lower the computational requirements of data sampling and processing. 

\section{Theoretical Analysis and Fingerprint Design}
\label{sec:theoretical-analysis}
Inspired by the mechanoreceptors in the human skin and the existing work on vibrotactile sensors and microphones for dynamic tactile sensing, we propose 3D-printed fingerprints to increase the body-borne vibrations and signal-to-noise ratio generated when interacting with objects and surfaces.
The vibrations are measured with contact microphones. Only one microphone is needed to differentiate material texture, while three or more microphones would allow for sound-source localization. The microphones can detect vibrations without a fingerprint pattern which we use as our baseline condition. 

\begin{figure}[htb]
  \centering
  \includegraphics[width=\columnwidth]{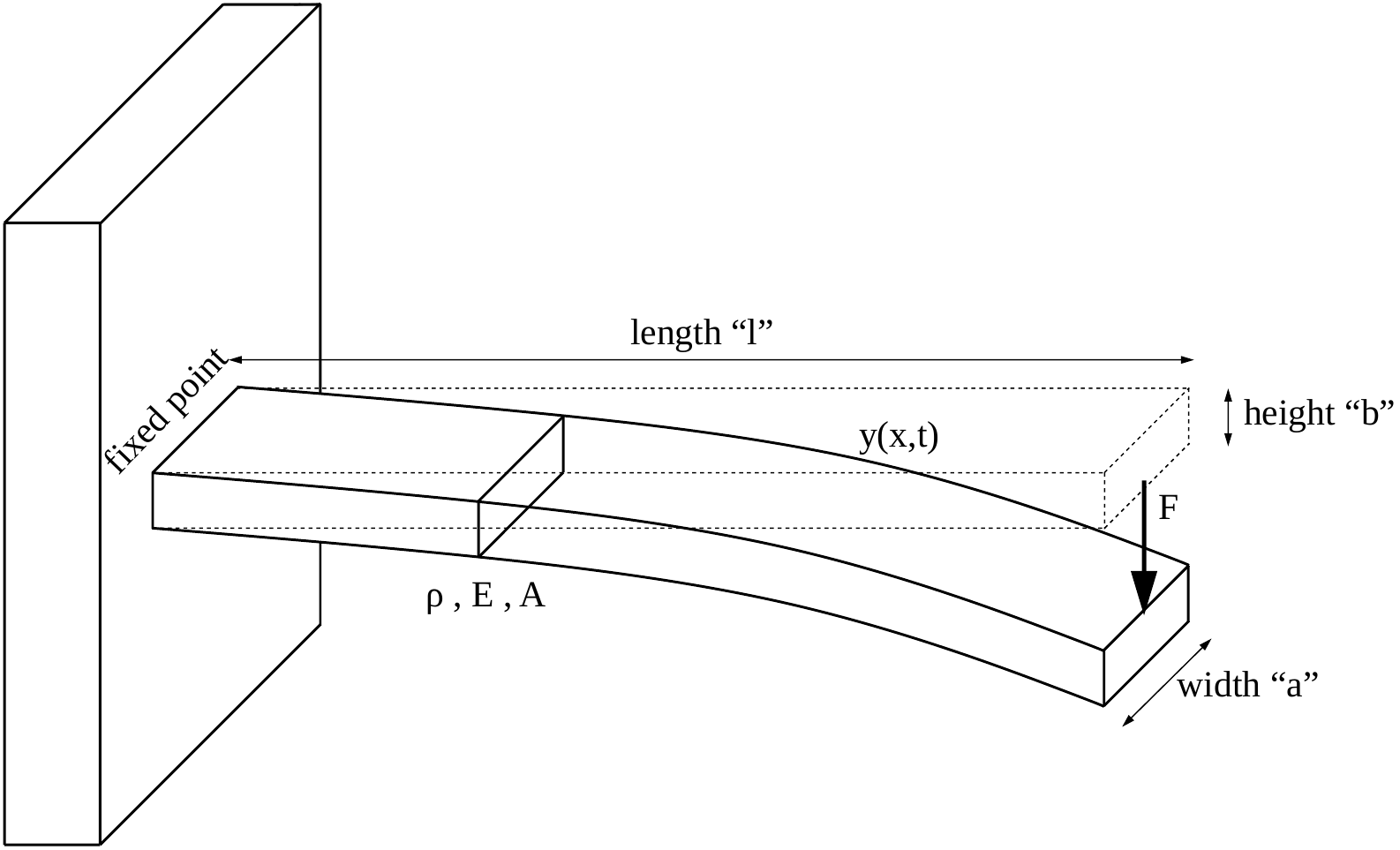}
  \caption{The beam is represented as a 3D prismatic solid of width $a$, height $b$ and length $l$. The deflection of the beam produced by a force $F$ depends on the Young's modulus $E$, the area moment of inertia $I$, the cross-sectional area $A$ and, the density of the material $\rho$.}
  \label{fig:fixed-free-beam}
\end{figure}

For the fingerprint pattern, we suggest an array of symmetric homogeneously spaced beams covering most of the robot's hand, as shown in Fig.~\ref{fig:RH8D-fingerprints}. We optimised the beams' length and cross-section to maximise the vibration signals' spectral magnitude within the
audible range. 
We used the Euler-Bernoulli beam equation for free vibration to determine the dimensions of a ``fixed-free'' beam (see Figure~\ref{fig:fixed-free-beam}) so that the beam's natural frequency is within the range of higher sensitivity of the microphones. 
Equation~\ref{eq:vibration-fixed-free-beam} represents the vibration of a beam \cite[p.\ 723]{Rao2010Mechanical}:
\begin{equation} 
  EI \pdv[4]{y}{x}(x,t) + \rho A \pdv[2]{y}{t}(x,t) = 0
  \label{eq:vibration-fixed-free-beam}
\end{equation}
\noindent where $y(x,t)$ represents the instantaneous position of a point within the beam at time $t$. $E$ is the Young's modulus of the material in $Pa$, $I$ is the area moment of inertia of the cross-section in $m^4$. $\rho$ the density of the material $\sfrac{kg}{m^3}$, $A$ the cross-sectional area in $m^2$.
The natural frequency $\omega$ of the beam in Hertz can be computed as

\begin{equation} 
  \omega_0 = \frac{1}{2\pi} (\beta_n l)^2 \sqrt{\frac{EI}{\rho A l^4}}
  \label{eq:fundamental-frequency}
\end{equation}
\noindent where $l$ is the length of the beam. $\beta_n$ represents the vibration modes and can be determined from the boundary conditions of the beam. In this study, we used $n=1$ for a ``fixed-free'' beam. Thus, $\beta_n l = 1.875104$ \cite[p.\ 726]{Rao2010Mechanical}.

\begin{figure}[htb]
  \centering
  \includegraphics[width=\columnwidth]{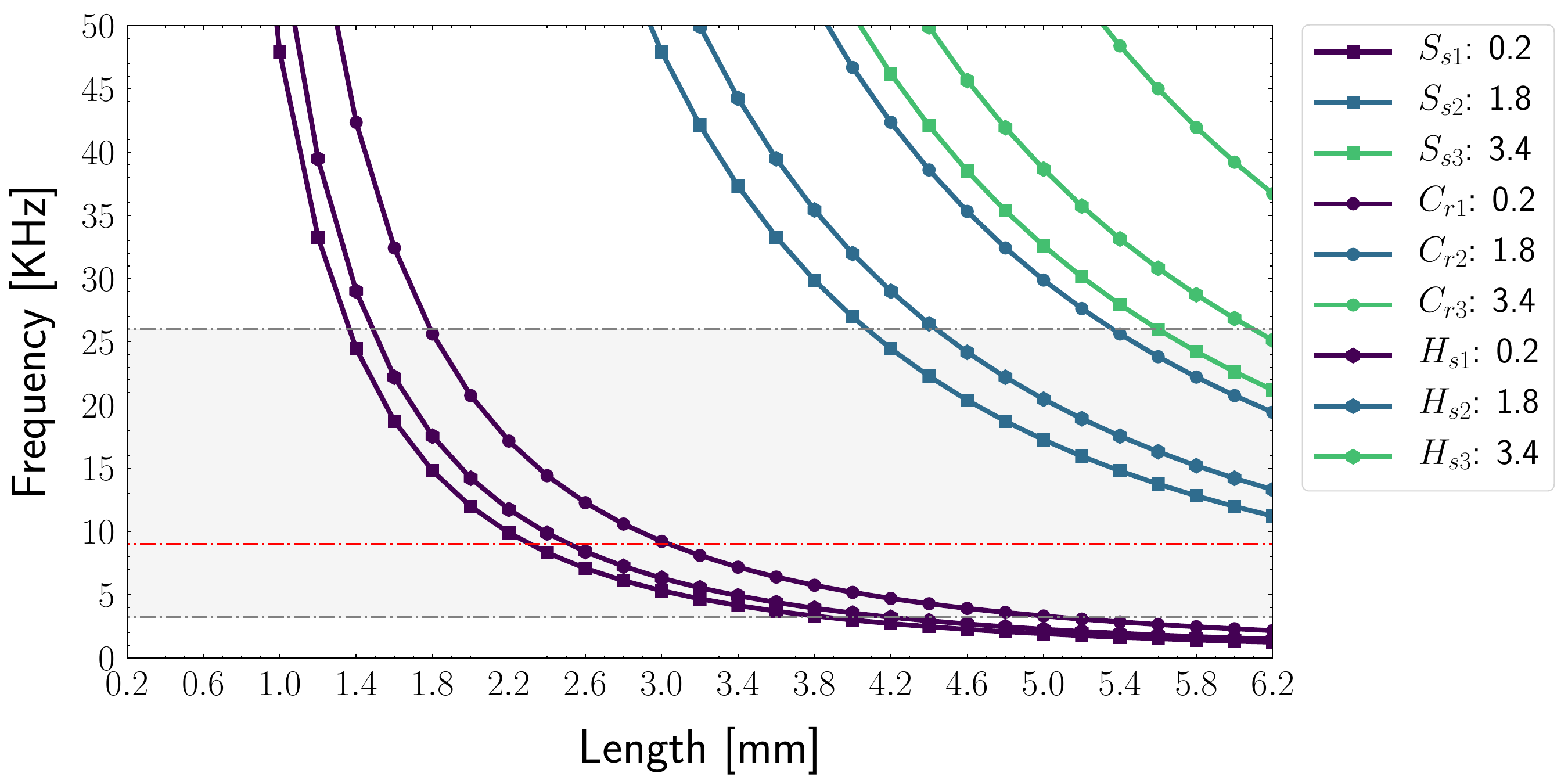}
  \caption{Natural Frequencies produced by different cross-sections: square ``$S_s$'', hexagon ``$H_s$'', and circle ``$C_r$''. The legend indicates the corresponding side and radius. The range of maximum sensitivity of the microphone is shown as a shaded area, and the peak at 9 kHz as a red line.}
  \label{fig:nf-beam-shapes-PLA}
\end{figure}

We analysed symmetric cross-section shapes for the beams: a square, hexagon, and circle. The square shape produces the lowest natural frequency, followed by the hexagon, while the circle leads to the highest natural frequencies for comparably sized beams, see Fig.~\ref{fig:nf-beam-shapes-PLA} for an example. 
From Equation~\ref{eq:fundamental-frequency}, it can also be inferred that solid beams lead to lower natural frequencies than hollow (lower density) beams. 

\begin{figure}[htb]
  \centering
  \includegraphics[width=\columnwidth]{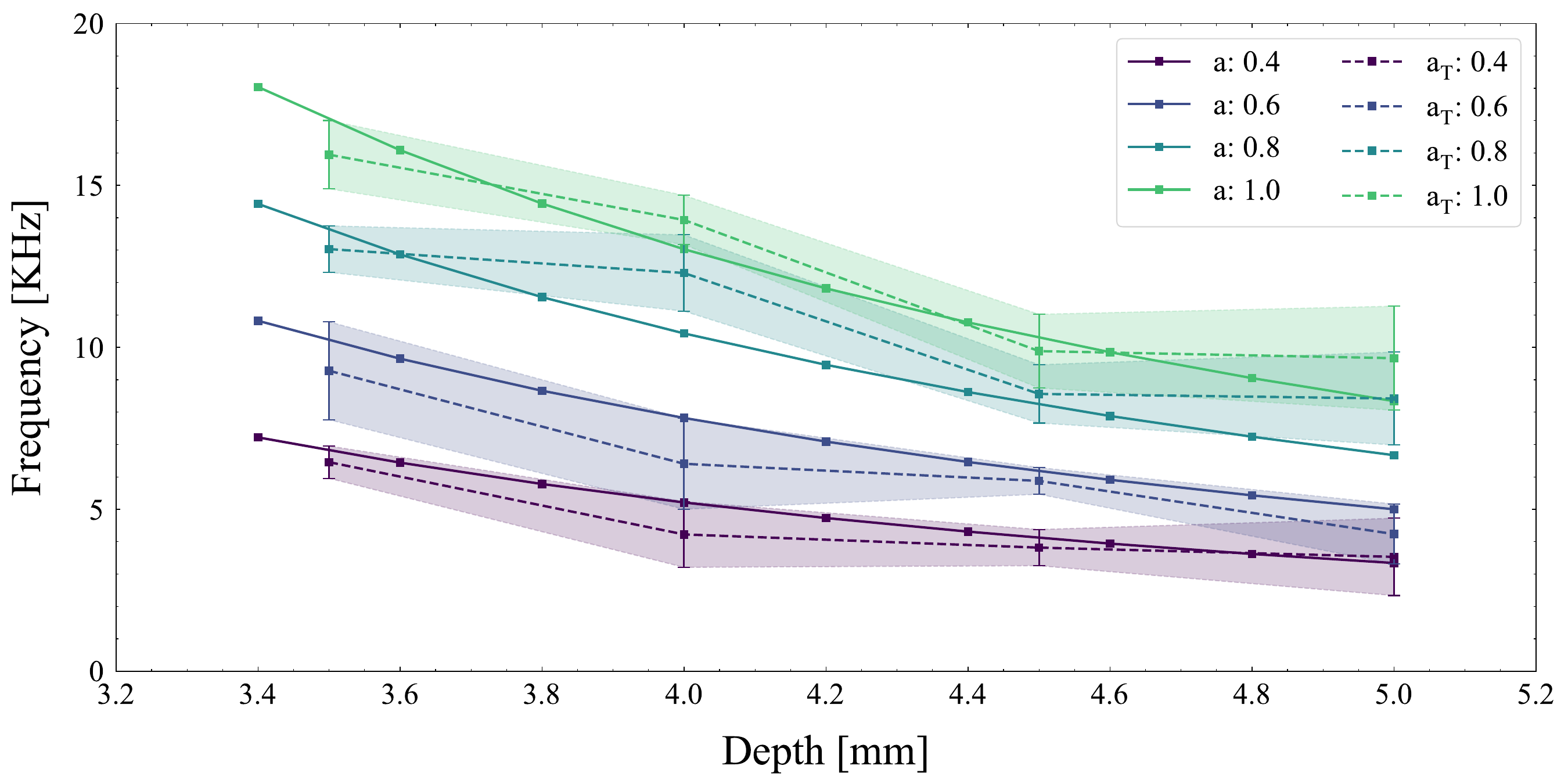}
  \caption{The natural frequency of 3D-printed square beams printed on ST 45B resin. The continuous lines represent the theoretical values ($a$), while the dotted lines ($a_T$) correspond to the mean of 10 measurements, and the shadowed area is the corresponding standard deviation. The legend indicates the corresponding side and radius.}
  \label{fig:test-single-beam-depth}
\end{figure}

We verified the analysis of the peak resulting frequency on a small sample of 3D-printed square beams printed on ST 45B resin. Figure~\ref{fig:test-single-beam-depth} shows the result of 10 samples for beams of different widths and lengths. The two smallest cross-sectional areas match the closest to the expected natural frequency.

\subsection{Final Design}
Based on the simulated behaviour and preliminary test for PLA, TPU, and ST 45B resin, we determined that the PLA and ST 45B resin behave similarly. Thus, the same design can be used for both materials. However, TPU is a soft material requiring a beam of larger cross-sectional area to keep the length and natural frequency low. 

A solid square beam seems the most suitable shape for the cross-section because it leads to a low natural frequency for the impulse response without increasing beam length in contrast to a cylindrical shape. Finally, considering the 3D printing guidelines [24] described in Section \ref{sec:requirements}, we determine the range for the side and length of a solid square beam, summarized in Table \ref{tab:beam-ranges}.

\begin{table}[htbp]
    \centering
    \caption{Ranges of length and side for the solid square beams of different 3D-printing materials.}
    \label{tab:beam-ranges}
    \begin{tabular}{lcc}
        \hline\noalign{\smallskip}
        Material & Side & Length \\
        \noalign{\smallskip}\hline\noalign{\smallskip}
        PLA or ST 45 B & [0.4, 1.0] & [3.4, 4.0]\\
        TPU or FL 300 & [2.0, 2.6] & [1.4, 2.0]\\
        \noalign{\smallskip}\hline
    \end{tabular}
\end{table}

An additional constraint to consider for the final design is the length of the beam not to restrict the motion of the fingers and still allow for the hand to close. The final dimensions of the different segment types used to collect the dataset are summarized in Table \ref{tab:final-beams}.

\begin{table}[htbp]
  \centering
  \caption{Dimensions of the beams for the final design.}
  \label{tab:final-beams}
  \begin{tabular}{lcc|cc}
    \hline\noalign{\smallskip}
    \multirow{2}{*}{Segment} & \multicolumn{2}{c|}{PLA or ST 45 B} & \multicolumn{2}{c}{TPU or FL 300} \\
    & Side & Length & Side & Length \\
    \noalign{\smallskip}\hline\noalign{\smallskip}
    Finger tip & 1.0 & 4.0 & 2.6 & 2.0 \\
    Finger phalanges & 1.0 & 3.5 & 2.6 & 1.8 \\
    Thumb phalanges & 1.0 & 3.2 & 2.6 & 1.6 \\
    Palm & 1.0 & 3.5 & 2.6 & 1.8 \\
    \noalign{\smallskip}\hline
  \end{tabular}
\end{table}

\begin{figure}[tb]
  \centering
  \includegraphics[height=45mm]{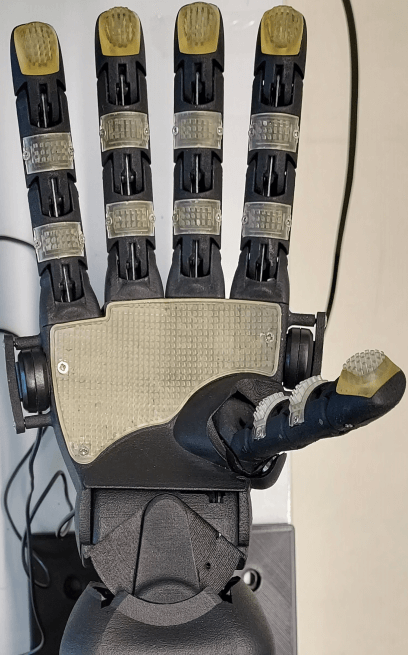}
  \includegraphics[height=45mm]{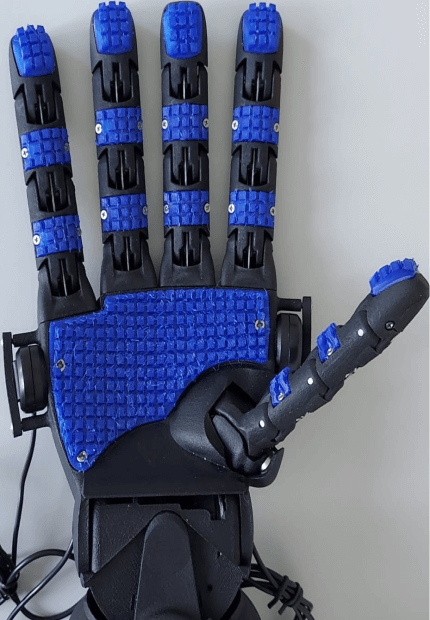}
  \caption{RH8D hand with fingerprint patterns. Left: ST 45B resin. Right: TPU. The CAD files, data and demonstration video have been made available to encourage reproducibility at \href{https://doi.org/10.6084/m9.figshare.21120982}{doi: 10.6084/m9.figshare.21120982}.}
  \label{fig:RH8D-real-fingerprints}
\end{figure}

The final designs are shown in Fig.~\ref{fig:RH8D-real-fingerprints}. The ST 45B resin design includes a small border for all the segments and is intended to minimize beams breaking under undesired interactions or extreme flexion. Those design features are more clearly visible in the render shown in Fig. \ref{fig:RH8D-fingerprints}. The beams are separated by a distance equal to the beam side. 

\section{Results}
\label{sec:results}
\begin{figure}[htb]
  \centering
  \includegraphics[width=0.87\columnwidth]{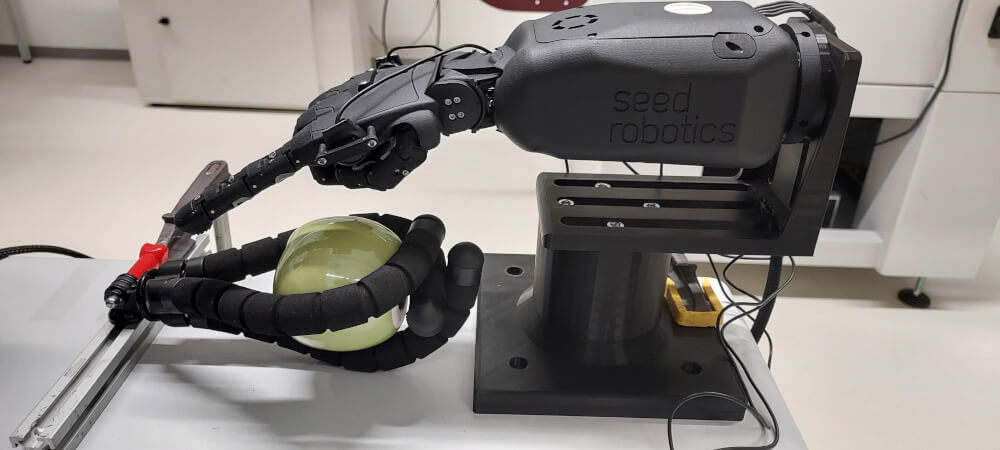}
  \caption{Experimental setup. The object is securely held in front and centre of the hand. The base of the object holder (marked as red) is 20cm away from the centre of the robot's palm.}
  \label{fig:lateral-motion-setup}
\end{figure}

We tested the designs on the real robot hand with five objects: porcelain apple, glass bottle, sponge, wooden cylinder, and steel beam. The test consisted of sliding the index finger over the object (Lateral Motion EP), see Fig.~\ref{fig:lateral-motion-setup}. 
Force control for the hand is performed using numerical 12 bits coding. Unfortunately, no correspondence to a force unit can be found in the documentation. For our experiments, we used a force setting of 400, which we determined to be the lowest setting for which friction could be overcome for all objects and fingerprint materials tested. 
The microphones recorded the vibrations with a sampling rate of $500 kHz$. 

\begin{figure}[htb]
  \centering
  \includegraphics[width=\columnwidth]{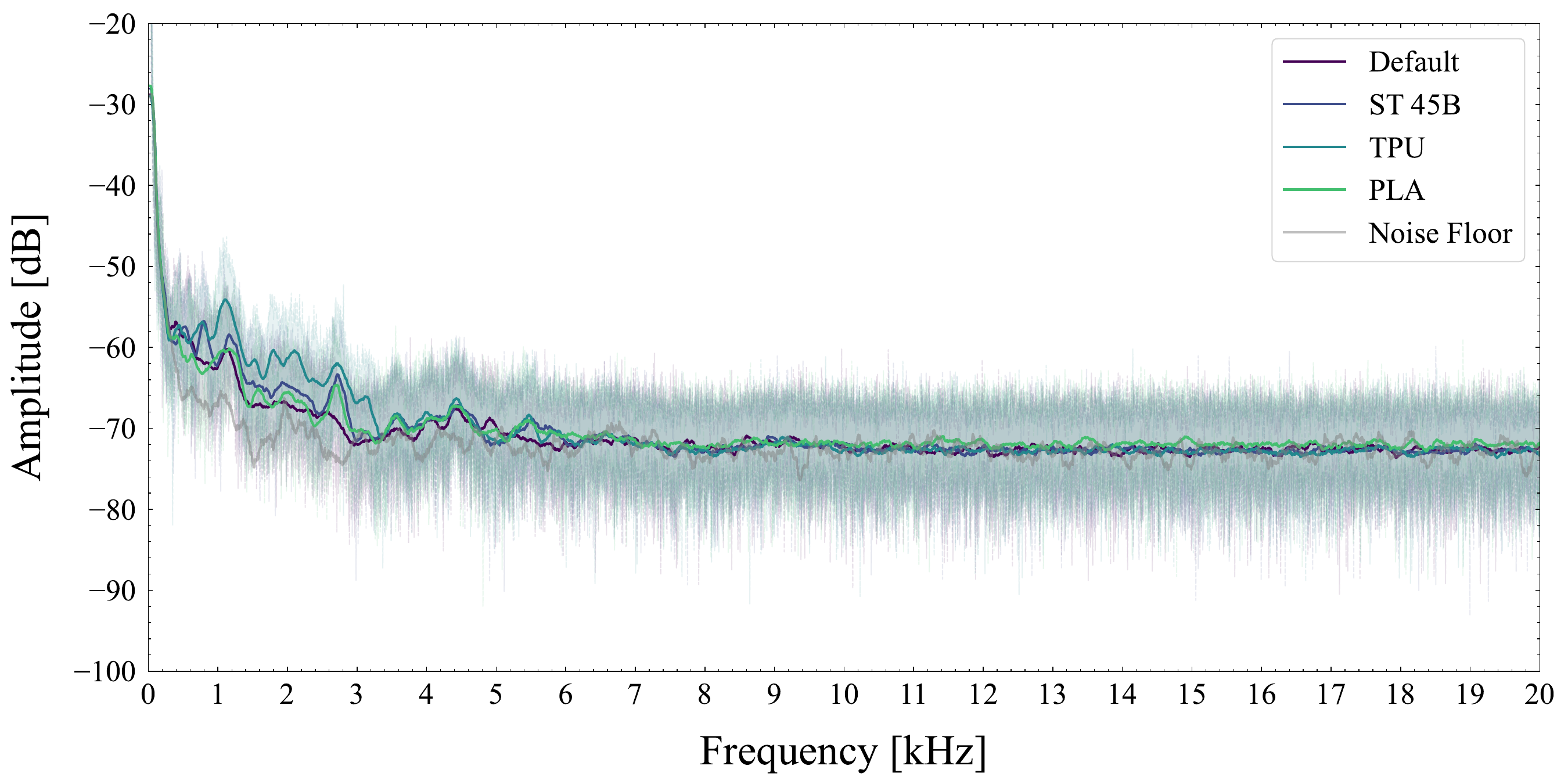}
  \caption{Mean frequency response of the ten Lateral Motion EP using the fingerprints on the porcelain apple measured at the left microphone.}
  \label{fig:lateral-motion-left-porcelain}
\end{figure}

An example of the mean frequency response for ten repetitions of the Lateral Motion EP on a porcelain apple is shown in Fig.~\ref{fig:lateral-motion-left-porcelain}. The optimized patterns lead to higher amplitude vibrations than the default ``robot skin'' (baseline) within the first $5kHz$.

\begin{figure}[htb]
  \centering
  \includegraphics[width=\columnwidth]{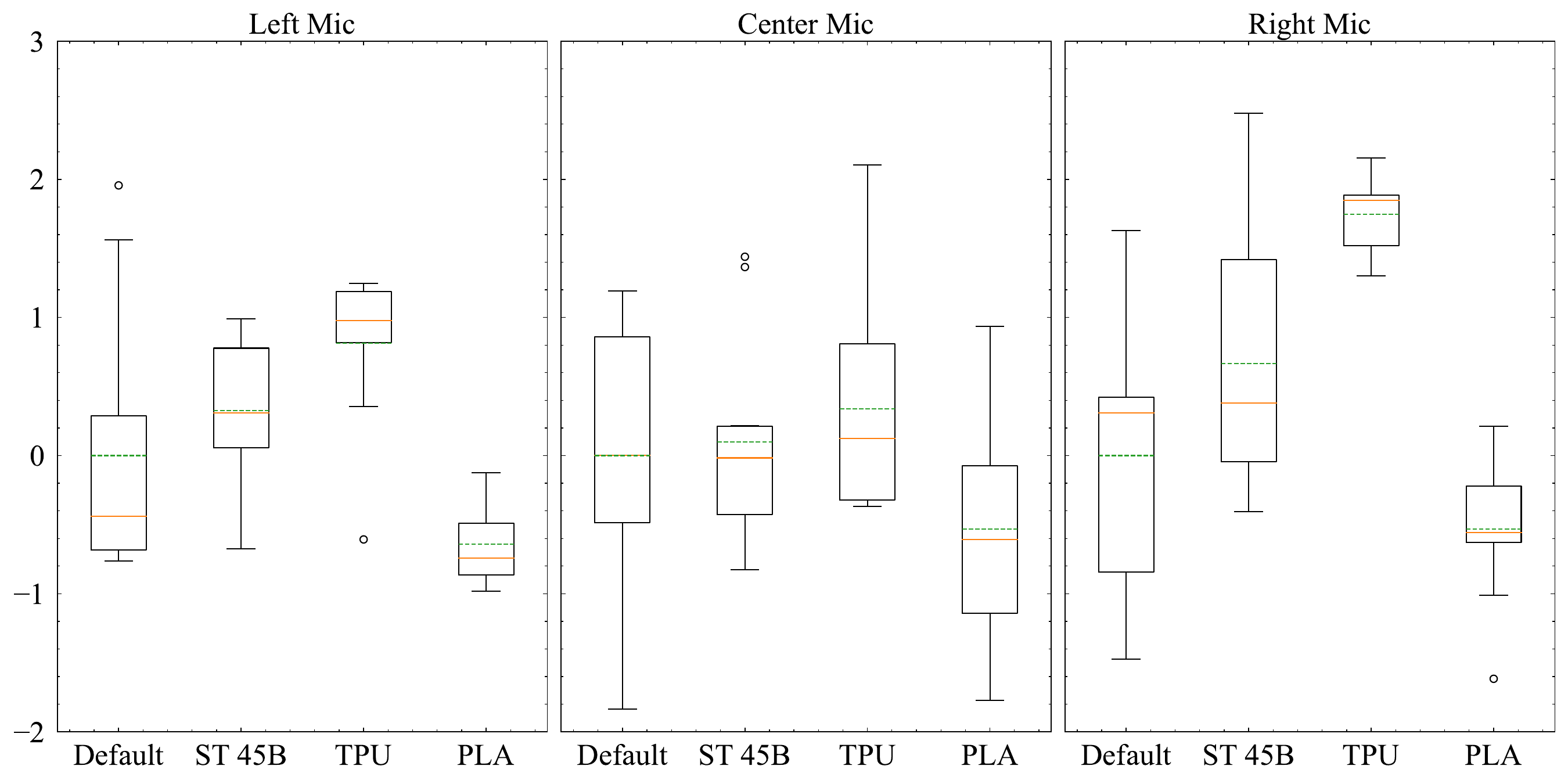}
  \caption{Comparison of Area Under the Curve of the frequency response produced by sliding the index finger over a \textit{sponge} 10 times. Default is the baseline using the commercial/original fingerprint segments. The results are normalized with respect to the baseline.}
  \label{fig:finger-sponge}
\end{figure}

\begin{figure}[htb]
  \centering
  \includegraphics[width=\columnwidth]{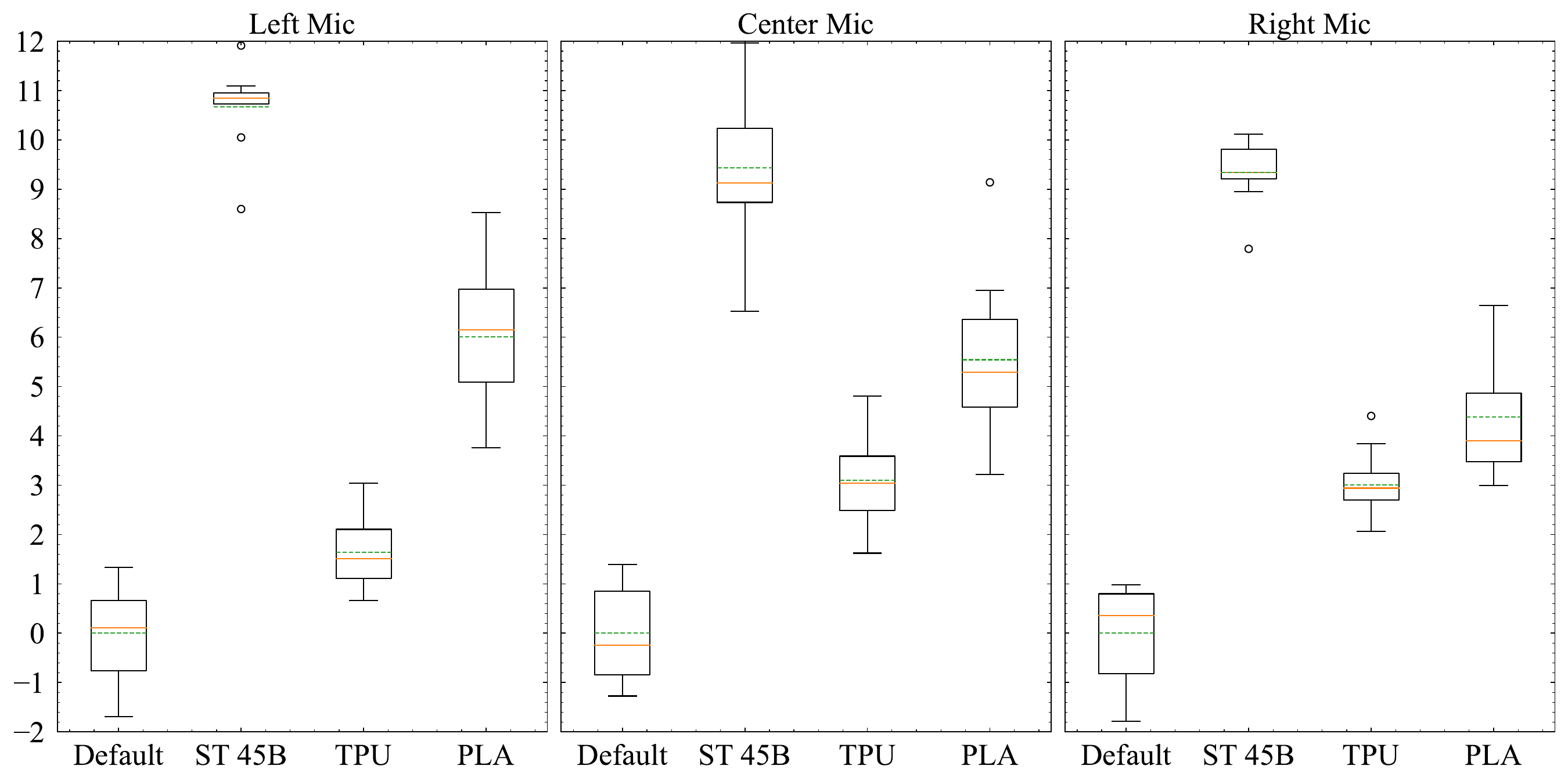}
  \caption{Comparison of Area Under the Curve sof the frequency response produced by sliding the index finger over a \textit{wooden stick} 10 times. Default is the baseline using the commercial/original fingerprint segments. The results are normalized with respect to the baseline.}
  \label{fig:finger-wood}
\end{figure}

The spectral amplitude diagram makes the quantification of the differences challenging. Thus, we computed the area under the curve for each test. The mean area for the baseline was used to normalize the areas of the different fingerprint materials such that in Fig.~\ref{fig:finger-sponge} and Fig.~\ref{fig:finger-wood}, the y-axis represents how many times the amplitude increased or decreased with respect to the baseline. Each microphone is normalized separately. 

When interacting with softer/deformable objects such as a sponge, the proposed fingerprints produce a comparable level of vibrations as the baseline, see Fig.~\ref{fig:finger-sponge}. Here the fingerprint of the more flexible material (TPU) leads to slightly higher vibrations than the harder fingerprints.

In contrast, when interacting with more rigid objects such as a wooden stick, an increase of over 11 times the baseline can be reliably measured, see Figure~\ref{fig:finger-wood}. In this case, harder fingerprint materials lead to considerably higher frequency response than both the baseline and the more flexible fingerprint material.

\subsection{Dataset}
\label{sec:dataset}
Based on our results, we used the ST 45B resin fingerprint patterns to create a vibrotactile dataset consisting of pictures of each object and haptic data. A total of 52 objects were recorded. 
Five observations per object were collected, and the objects were reoriented each time. The haptic data consist of the vibrations generated by the fingerprint patterns, motor position information and current values.
The object explorations include four object exploration procedures. Rubbing the object (Lateral Motion EP) with the middle and index finger with a force of 400 (12 bits coding), enclosing the object using a force of 300 (12 bits coding) for 2 seconds (Enclosure EP), and squeezing the object with a force of 400, 500, 600, and 700 (Pressure EP). Finally, the current delivered to the wrist is recorded (Unsupported Holding EP). 
The dataset, CAD files and demonstration videos can be found under \href{https://doi.org/10.6084/m9.figshare.21120982}{doi: 10.6084/m9.figshare.21120982}.

\section{Discussion}
\label{sec:discussion}
The density and Young's modulus are proportional to the generated frequencies. Similarly, more flexible materials lead to lower frequencies. 

During our experiments, we noted that rigid fingerprints slid over the objects with hard surfaces, thus generating low vibrations. Using materials with a higher friction coefficient would be recommended in such cases. Alternatively, anisotropic patterns could be used to cope with a more extensive variety of interactions because no fingerprint material performed best across all objects. 





The mathematical modelling of a single beam can be verified in the 3D-printed counterpart.
However, the measured frequencies on the real robot are lower than the calculated ones. 
The discrepancies between the simulated and actual measurements can partly be attributed to the theoretical analysis performed for a single beam and a single perturbation (impulse response). The hand's kinematics should be considered to improve the design further. 
Moreover, considering that the obtained frequencies were lower than expected would allow for a shallower fingerprint pattern, which should be more robust. The dimensions of the beam were, in part, chosen to produce low frequencies.

\section{Conclusions}
\label{sec:conclusions}
Vibrotactile sensing is a sound approach to address some of the challenges of tactile sensing. 
The strategy has been tested in various applications, including tactile sensing, texture perception for prosthetics and social touch classification, as discussed in Section \ref{sec:sota}. 
However, no systematic approach for designing such artificial fingerprints has been suggested yet. 

In this article, we suggested an approach to optimizing 3D-printed fingerprint patterns based on the Euler-Bernoulli beam equation. 
The 3D-printed fingerprint material and geometrical dimensions were analytically determined to maximize the vibrations generated within the microphone's sensitive range. 
The optimized fingerprints can produce over 11 times higher amplitude than the baseline. 

Based on our results, the ST 45B resin achieves a higher frequency response across different object materials. Thus, this fingerprint design and material were used to create a public vibrotactile dataset including 52 objects.

Although detecting body-borne vibrations are limited to dynamic interaction and cannot provide force information, they can be complemented by kinesthetic information to estimate force or other tactile sensing modalities. 
Despite these disadvantages, our 3D-printed fingerprints approach exploiting body-borne vibrations as tactile sensing offer several benefits. For instance, it is easy to manufacture, can be scalable, and it can be designed for curved surfaces. Additionally, the sensor is robust since the active sensing elements can be mounted inside the robot.

Future research would also include studying the effect of fingerprint patterns of different materials, shapes and separations to more effectively deal with a wide range of object materials, such as rigid or deformable objects. 

In the future, we would also like to study the effects of fingerprint wear, the effect of interaction force, velocity and angle of incidence of the fingerprint pattern, as all these variables could impact the resulting vibration pattern. 

Another line of future research includes sound-source localization on the robot's body to infer the single or potential multi-contact points. The provided dataset could be used to explore this line of research.

%

\addtolength{\textheight}{-149mm}   



%
\section*{Acknowledgement}
This work was supported by the European Commission under the Horizon 2020 framework program for Research and Innovation via the APRIL project (project number: 870142) and the VeryHuman project funded by the \href{http://www.bmbf.de/en/}{Federal Ministry of Education and Research} with grant no.\ 01IW20004.

\bibliographystyle{IEEEtran}
\bibliography{IEEEabrv,references.bib}

\end{document}